\def\year{2022}\relax
\documentclass[letterpaper]{article} 
\usepackage{aaai22}  
\usepackage{times}  
\usepackage{helvet}  
\usepackage{courier}  
\usepackage[hyphens]{url}  
\usepackage{graphicx} 
\usepackage{natbib}  
\usepackage{caption} 
\DeclareCaptionStyle{ruled}{labelfont=normalfont,labelsep=colon,strut=off} 
\usepackage{algorithm}
\usepackage{algorithmic}

\usepackage{newfloat}
\usepackage{listings}
\floatstyle{ruled}
\newfloat{listing}{tb}{lst}{}
\floatname{listing}{Listing}

\usepackage{booktabs}
\usepackage{comment}
\usepackage{amssymb}
\usepackage{pifont}
\newcommand{\cmark}{\ding{51}}%
\newcommand{\xmark}{\ding{55}}%
\title{Cross-Dataset Collaborative Learning for Semantic Segmentation \\ in Autonomous Driving}

\author{Li Wang \textsuperscript{\rm 1}, Dong Li \textsuperscript{\rm 1}, Han Liu \textsuperscript{\rm 1}, Jinzhang Peng \textsuperscript{\rm 1}, Lu Tian \textsuperscript{\rm 1}, Yi Shan \textsuperscript{\rm 1}\\
\textsuperscript{\rm 1} Xilinx Inc. \\
\tt\small{\{liwa, dongl, hanl, jinzhang, lutian, yishan\}@xilinx.com}\\
}

\usepackage{bibentry}

\begin{document}

\maketitle

\begin{abstract}
Semantic segmentation is an important task for scene understanding in self-driving cars and robotics, which aims to assign dense labels for all pixels in the image. Existing work typically improves semantic segmentation performance by exploring different network architectures on a target dataset. Little attention has been paid to build a unified system by simultaneously learning from multiple datasets due to the inherent distribution shift across different datasets. In this paper, we propose a simple, flexible, and general method for semantic segmentation, termed Cross-Dataset Collaborative Learning (CDCL). Our goal is to train a unified model for improving the performance in each dataset by leveraging information from all the datasets. Specifically, we first introduce a family of Dataset-Aware Blocks (DAB) as the fundamental computing units of the network, which help capture homogeneous convolutional representations and heterogeneous statistics across different datasets. Second, we present a Dataset Alternation Training (DAT) mechanism to facilitate the collaborative optimization procedure. We conduct extensive evaluations on diverse semantic segmentation datasets for autonomous driving. Experiments demonstrate that our method consistently achieves notable improvements over prior single-dataset and cross-dataset training methods without introducing extra FLOPs. Particularly, with the same architecture of PSPNet (ResNet-18), our method outperforms the single-dataset baseline by 5.65\%, 6.57\%, and 5.79\% mIoU on the validation sets of Cityscapes, BDD100K, CamVid, respectively. We also apply CDCL for point cloud 3D semantic segmentation and achieve improved performance, which further validates the superiority and generality of our method. Code and models will be released.
\end{abstract}

\section{Introduction}
\begin{figure}[t!]
\footnotesize
\begin{center}
\begin{tabular}{@{}cc@{}}
\includegraphics[width = 0.4\linewidth]{{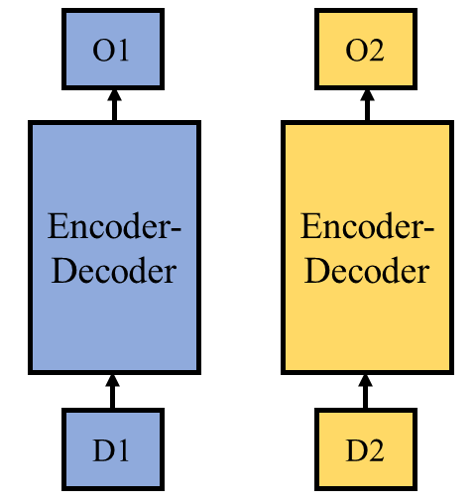}} & 
\includegraphics[width = 0.4\linewidth]{{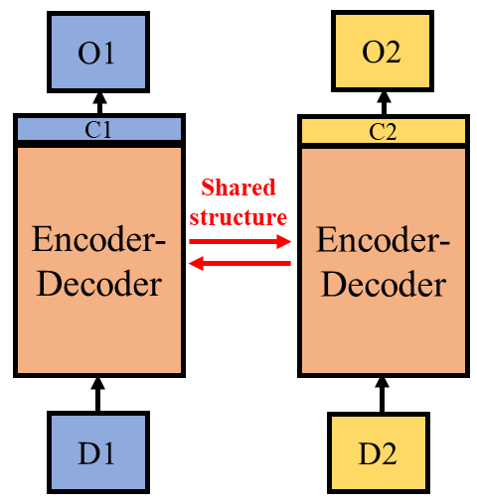}} \\
{\footnotesize (a) Single-Dataset Training} &
{\footnotesize (b) Finetuning}
\end{tabular}
\begin{tabular}{@{}cc@{}}
\includegraphics[width = 0.4\linewidth]{{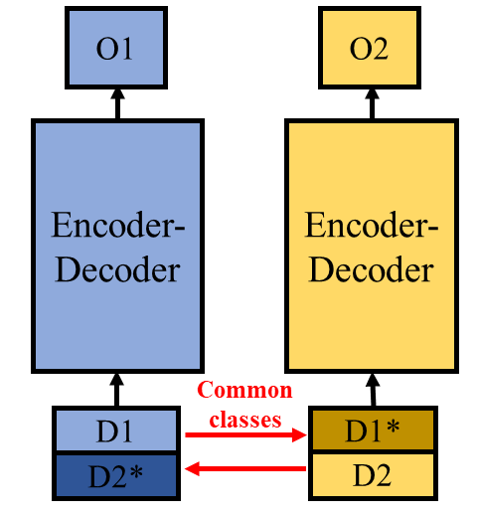}} & 
\includegraphics[width = 0.4\linewidth]{{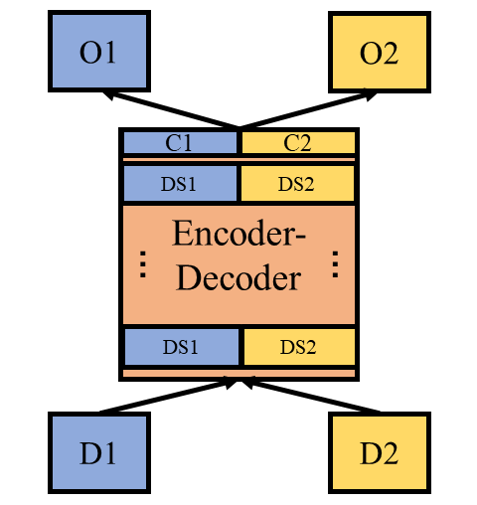}}\\
{\footnotesize (c) Label Remapping} &
{\footnotesize (d) CDCL (Ours)}
\end{tabular}
\end{center}
\caption{Illustration of prior baselines and the proposed algorithm in the cross-dataset setting. (a) Single-dataset training: each network is trained on each dataset separately. (b) Finetuning: each network is pretrained on source and finetuned on target with shared structure. (c) Label remapping: each network is trained on the combination of the target dataset and a subset of the source dataset with common classes. (d) CDCL (Ours): a unified network is trained on all the datasets with data-aware block (DAB) and dataset alternation training (DAT). Notation: D: Dataset, O: Output, C: Classifier layer, DS: Dataset-Specific layer.}
\label{figure: idea}
\end{figure}

Semantic segmentation is a fundamental task in scene understanding, which aims to partition the image into semantically meaningful parts. It has been widely applied to practical vision-based systems, such as self-driving cars and robotics, medical aided diagnosis and augmented reality.

Owing to the development of deep learning, semantic segmentation has achieved remarkable improvements in recent years. The success is mainly attributed to discriminative features learned from network backbone, tailored encoder-decoder framework and dense annotations for supervised learning. Various deep networks and their variants (e.g., AlexNet ((Krizhevsky et al. 2017), ResNet \cite{he2016deep}) provide discriminative features to recognize semantics in images. Different encoder-decoder frameworks are developed for semantic segmentation. Especially, the mIoU performance on the Cityscapes benchmark \cite{cordts2016cityscapes} is improved from 65.3\% by FCN (Long et al. 2015) to 82.7\% by DeepLabv3+ \cite{chen2018encoder}. Moreover, multiple semantic segmentation datasets blossom (e.g., Cityscapes \cite{cordts2016cityscapes}, BDD100K \cite{yu2018bdd100k}), which provide rich labeled data for network training. However, how to build a unified system by simultaneously training from several datasets has not been well investigated by the community. In this work, we consider cross-dataset semantic segmentation and expect to learn a general model from multiple datasets. Such cross-dataset setting is useful for practical applications. For example, in the autonomous driving scenarios, each dataset can be collected from different scenes, weather and illumination. Learning a unified model may exploit information from several datasets and help improve each other's performance, particularly for those with limited data.

Figure \ref{figure: idea} illustrates prior single-dataset and cross-dataset baseline methods. Most of existing semantic segmentation methods follow the single-dataset setting, i.e., the model is trained and tested on a single dataset only (Figure \ref{figure: idea} (a)). When directly apply the model to another dataset, it tends to yield inferior performance because of the distribution gap between the two datasets. A straightforward approach to improve the performance on the target dataset is finetuning the pretrained model with shared structure except the final prediction layer (Figure \ref{figure: idea} (b)). However, finetuning requires extra training epochs on target and may lead to degraded performance on source. It is not a unified model for maintaining high performance for both source and target datasets. Another cross-dataset solution is to exploit image samples with common classes from other datasets so as to enrich the training set (Figure \ref{figure: idea} (c)). Although training with more data sometimes can boost the performance, it counts on the assumption of class overlap among different datasets and requires carefully remapping the labels. Moreover, it often encounters inconsistent taxonomies and annotation criteria, leading to limited application.

To alleviate these problems, we propose a Cross-Dataset Collaborative Learning (CDCL) algorithm that is capable of learning from multiple datasets (Figure \ref{figure: idea} (d)). First, we present a sightful investigation on how a model differs by training on different datasets separately. Our key observations lie in: (1) The convolution (Conv) filters can be shared for all datasets to maintain network efficiency without accuracy loss. (2) The batch normalization (BN) layers are not appropriate to share across different datasets due to the bias of statistics. Second, motivated by the observations, we present a unified network to preserve the commonality and particularity of different datasets. Specifically, we introduce a Dataset-Aware Block (DAB) as the fundamental computing unit of the network, which helps capture homogeneous convolutional representations and heterogeneous statistics across different datasets. The proposed block is composed of a dataset-invariant convolution layer, multiple dataset-specific batch normalization layers, and an activation layer. Moreover, we propose a Dataset Alternation Training (DAT) mechanism to facilitate the collaborative optimization procedure. Our method is simple, easy-to-implement, and compatible with mainstream semantic segmentation frameworks. As the Conv layers are shared across different datasets, our network does not introduce extra computation cost compared to the single-dataset baseline\footnote{The number of BN parameters is far less than that of Conv weights in our networks.}. We conduct extensive experiments on diverse semantic segmentation datasets for autonomous driving. Experimental results demonstrate our method consistently outperforms prior single-dataset and cross-dataset training methods. Particularly, with the same architecture of PSPNet (ResNet-18), our method surpasses the single-dataset baseline by 5.65\%, 6.57\%, and 5.79\% mIoU on the validation sets of Cityscapes, BDD100K, CamVid, respectively.

The main contributions of this paper are summarized as follows. (1) We analyze the limitations of existing single-dataset and cross-dataset semantic segmentation methods, and provide insights on how a model differs by training on different datasets separately. (2) We propose a simple, flexible, and general cross-dataset collaborative learning algorithm, which can alleviate the distribution shift problem in the training phase and introduce no extra computation cost in the inference phase for each dataset. (3) We demonstrate the effectiveness of the proposed approach on diverse semantic segmentation datasets for autonomous driving. Our method consistently outperforms prior single-dataset and cross-dataset training baselines with the same computation budget. We also achieve improved performance for point cloud 3D semantic segmentation, which further validates the superiority and generality of our method.

\section{Related Work}
\subsection{Semantic Segmentation}
Semantic segmentation is a dense image prediction task, which plays a key role in high-level scene understanding. FCN \cite{long2015fully} and its follow-ups \cite{zhao2017pyramid,chen2014semantic,chen2017deeplab,chen2018encoder} have achieved impressive performance for semantic segmentation. Recent transformer-based methods \cite{zheng2021rethinking,xie2021segformer,liu2021swin,strudel2021segmenter} have also shown promising results on semantic segmentation benchmarks with self-attention architectures. In addition, crucial strategies have been developed to further improve the performance, including atrous convolution \cite{chen2017rethinking}, pyramid pooling module \cite{zhao2017pyramid}, attention mechanism (Hu et al. 2018; Fu et al. 2019b) and context encoding \cite{zhang2018context}. In parallel, light-weight networks \cite{mehta2018espnet,mehta2019espnetv2,paszke2016enet} arouse great research interest owing to their high speed and wide applications on resource-constrained devices. However, these semantic segmentation methods typically follow the setting of single-dataset training and can not maintain high accuracy on other datasets without finetuning.

\subsection{Cross-Dataset Training}
Recurrent Assistance (Perrett et al. 2017) first proposed the cross-dataset training mechanism, which is used for frame-based action recognition during the pretraining stage. Recent work explores cross-dataset training for object detection \cite{yao2020cross,wang2019towards}. The work of \cite{yao2020cross} proposed to generate a hybrid dataset by simple label concatenation or label mapping since the number and identity of classes are different for each dataset. Inspired by Squeeze-and-Excitation (Hu et al. 2018), the work of \cite{wang2019towards} introduced a domain attention module to activate a single network for universal object detection tasks. Our work is a vanguard in cross-dataset semantic segmentation, which aims to learn a unified model by simultaneously training from multiple datasets.

\subsection{Transfer Learning}
Both of transfer learning and our cross-dataset training methods address the distribution shift problem between different datasets. The main differences are summarized below. (1) Domain adaptation (DA) methods aim to adapt a model from the source domain with adequate labeled data to the known target domain (i.e., training images are available) with little or no labeled data (Argyriou et al. 2016; Li et al. 2017; Gkioxari et al. 2014). (2) Domain generalization (DG) emphasizes generalization on the unknown domain (i.e., training images are not available). Different from DA and DG tasks, we focus on learning a unified model from multiple datasets simultaneously
and improving the performance of all known datasets.

\begin{figure}[t]
\footnotesize
\begin{center}
   \includegraphics[width=1.0\linewidth]{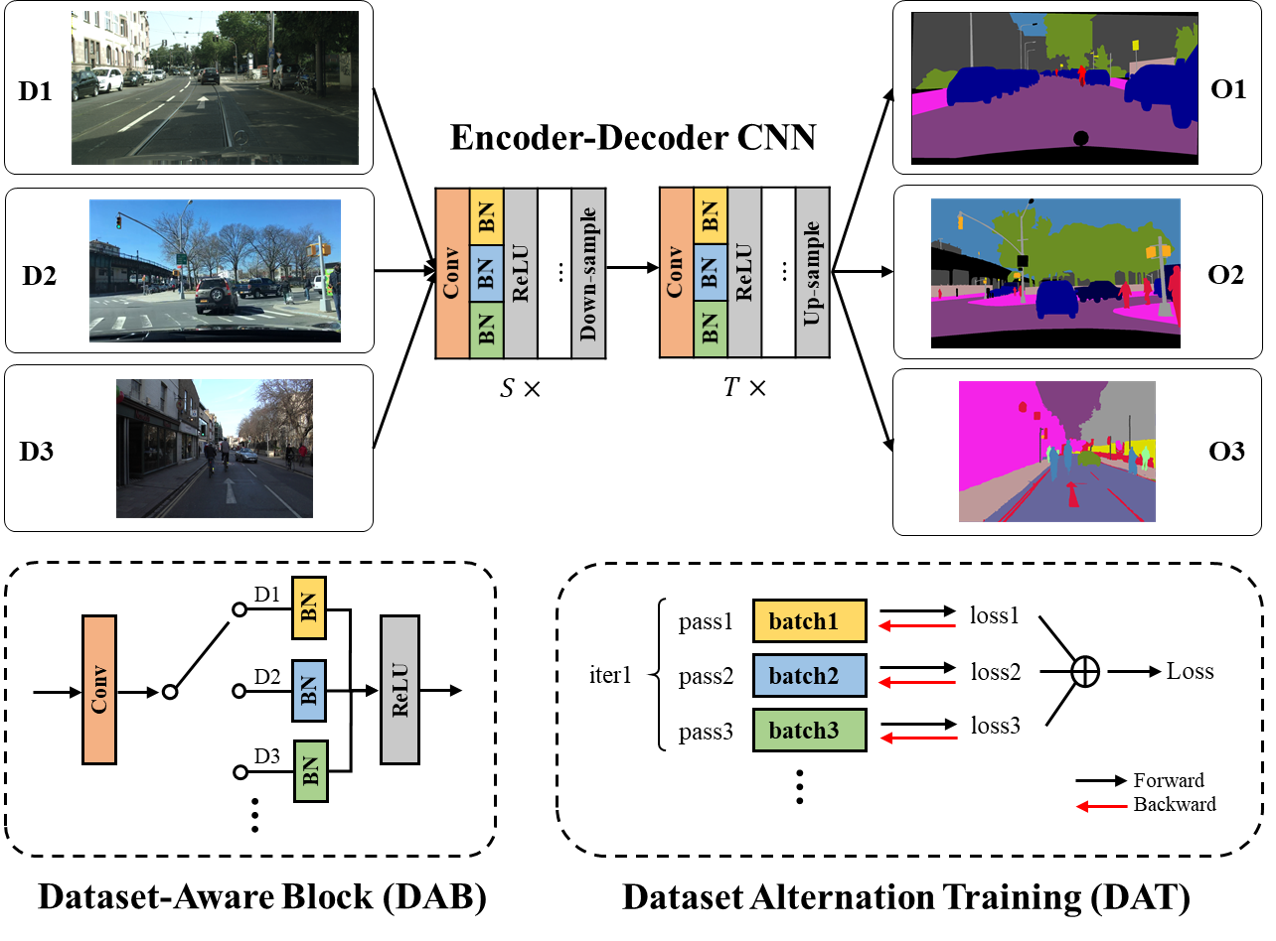}
\end{center}
    \caption{An overview of the proposed cross-dataset collaborative learning (CDCL) method. See Section \ref{sec:cdcl} for details.}
\label{fig:pipeline}
\end{figure}

\subsection{Multi-Task Learning}
Another related research topic is multi-task learning (MTL) \cite{argyriou2006multi,li2017integrated,gkioxari2014r}. The work of \cite{argyriou2006multi} jointly trained classification, detection, and segmentation tasks on a single dataset. It requires annotations for all tasks on a single dataset. The work of \cite{li2017integrated} trained an integrated face analysis model (facial landmark, facial emotion, and face parsing) by using multiple datasets, where each dataset is only labeled for one task. MTL methods can boost the performance by explicitly modeling the interaction of different tasks. In this paper, we aim to jointly train multiple datasets for a similar task. Furthermore, we analyze the commonality and particularity of training on different datasets and build an alternation training method for effective collaborative optimization.

\begin{figure}[t!]
\footnotesize
\begin{center}
   \includegraphics[width=1.0\linewidth]{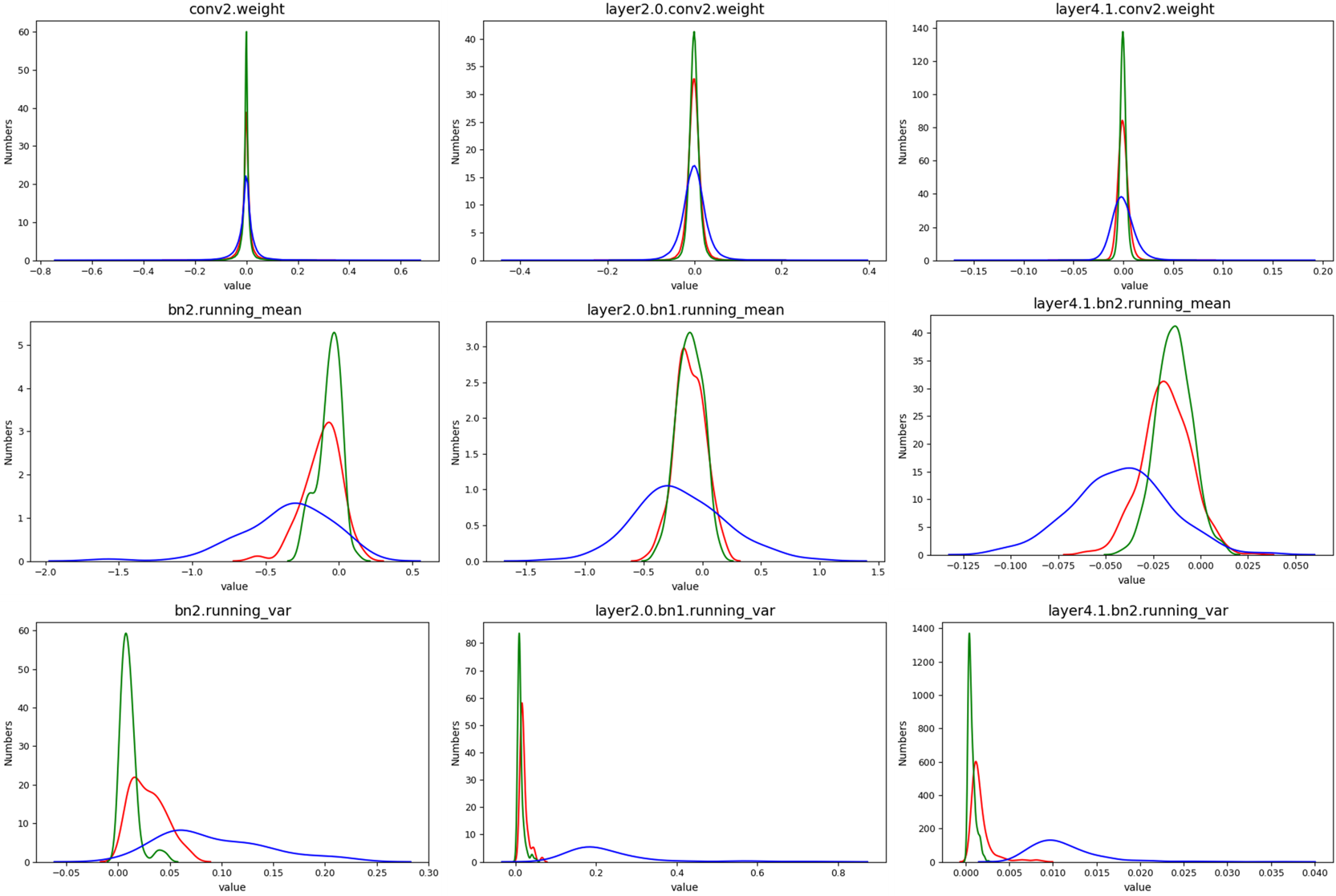}
\end{center}
\caption{Parameter distributions of Conv and BN layers on three datasets (red: Cityscapes, green: BDD100K, blue: CamVid). The networks are trained with PSPNet (ResNet-18) on each dataset separately. We sample three layers (left: \texttt{conv2} \& \texttt{bn2}, middle: \texttt{layer2}, right: \texttt{layer4}) and provide more curves of layers in the supplementary material.}
\label{fig:conv_bn_ana}
\end{figure}

\section{Approach}

In this section, we first analyze the most popular single-dataset and cross-dataset solutions for semantic segmentation. We then explain the motivation of the proposed cross-dataset collaborative learning algorithm and introduce each algorithmic component in detail.

\subsection{Analysis of Semantic Segmentation Methods}
\subsubsection{Single-Dataset Baseline.}
The encoder-decoder is used as the most popular architecture for semantic segmentation. It employs deep convolutional neural networks as the encoder to extract hierarchical features, and exploits a simple sub-network as the decoder to refine the segmentation results. For segmentation on a single dataset, recent work (e.g., PSPNet \cite{zhao2017pyramid} and DeepLab \cite{chen2018encoder}) has extended this basic architecture to improve the performance. For segmentation on multiple datasets, suppose we have $N$ datasets in total: $\{D_i\}_{i=1}^N$, the simplest solution is to train and evaluate a network for each dataset separately. This single-dataset baseline is inefficient since it needs to save $N$ sets of network parameters that are not shared. Furthermore, each network is only responsible for its corresponding dataset and the distribution shift between different datasets is not considered.

\subsubsection{Cross-Dataset Baseline.} Finetuning is a straightforward solution to handle the distribution shift problem and improve the accuracy on the target dataset. Specifically, it first pretrains the network on $D_i$ and then finetunes it on $D_j$. The finetuning baseline often needs to adjust hyper-parameters and requires extra training epochs to reach a satisfactory accuracy on the target dataset. In addition, it may lead to degraded performance on the source dataset as it does not incorporate source samples for joint optimization. Another solution is to exploit image samples with common classes from other datasets to enrich the training set of target dataset. For example, we denote $D_i^*$ as a subset of $D_i$ by remapping the common labels from $D_i$ to $D_j$. Training on combination of $D_i^*$ and $D_j$ may lift the performance owing to the usage of more data. However, such label remapping baseline has three limitations. (1) It counts on the assumption of class overlap among different datasets and not applicable to the case where datasets have disjoint classes. (2) It requires carefully remapping the labels when encountering inconsistent taxonomies and annotation criteria. For example, ``drivable'' and ``non-drivable'' areas are separately labeled in $D_j$ but only ``road'' is labeled in $D_i$, which incurs class conflict. (3) Naive combination of training samples from different datasets is likely to yield poor performance due to the discrepancy of image compositions, illuminations, scenes, etc.

\subsection{Cross-Dataset Collaborative Learning}
\label{sec:cdcl}
\subsubsection{Rethinking Convolution and Batch Normalization.}
To explore how to effectively alleviate the above dilemmas, we analyze the distribution shift problem across different semantic segmentation datasets. Intuitively, images from different datasets may vary greatly. Even though images come from a similar autonomous driving scenario, there exists inevitable appearance discrepancies as they are usually collected under different conditions (e.g., scenes and illuminations) in practice. That makes challenging to directly combine these images for joint training. Furthermore, we analyze the parameter distributions of Conv and BN layers when training on different datasets separately. Figure \ref{fig:conv_bn_ana} presents these parameter distributions of sampled low-/middle-/high-level layer on three different autonomous driving datasets (i.e., Cityscapes, BDD100K and CamVid). We summarize two key observations. (1) Conv weights hold similar distributions and most values are concentrated near zero. From the perspective of perception, each neuron in Conv layers only attends a local region. Such local information is less different than global appearance, which implies that Conv weights can be shared during the optimization of different datasets. (2) Both running mean and running variance of BN hold different distributions for different datasets, even for those sharing the same label space (e.g., Cityscapes and BDD100K). This is because the statistical moments (mean and variance) used in BN are relevant to each specific dataset. In detail, during training, the BN layer first calculates the mean and variance of the activations through the exponential moving average method (Ioffe et al. 2015), and then normalizes activations and applies a linear transformation to derive the layer's output. During inference, BN uses the estimated mean and variance for normalization and uses the learned linear transformation parameters to recover the representation ability of activations. We assume that the mean and variance used in BN compute the global statistics of a specific dataset, and thus both the normalization and linear transformation operations are sensitive to the dataset itself. These observations motivate us to use shared Conv and independent BN parameters for joint training on different datasets.

\subsubsection{Dataset-Aware Block.}
Inspired by the above analysis, we propose a simple and flexible framework for training on different semantic segmentation datasets simultaneously. As shown in Figure \ref{fig:pipeline}, we design a dataset-aware block (DAB) as the fundamental computing unit of the encoder-decoder architecture. Specifically, DAB consists of a dataset-invariant Conv layer, multiple dataset-specific BN layers, and an activation layer.
The weights of dataset-invariant Conv layers are shared, while the parameters of each dataset-specific BN layer are not shared across different datasets.
In each block, we learn $N$ BN layers for $N$ different datasets and each BN layer is responsible for a specific dataset. A switch is automatically to determine which BN should be activated based on the data source.
Therefore, our dataset-specific BN can be formulated by:

\begin{equation}
\mbox{BN}\{D_i\}(X_i;\gamma_i,\beta_i) = \gamma_i {\bar{X_i}} + \beta_i \\
\label{eq:dsbn}
\end{equation}
where,
\begin{equation}
\bar{X_i} = \frac{X_i - \mu_i}{\sqrt{\sigma_i^{2} + \epsilon}} \\
\mu_i = \frac{1}{B}\sum_{j=1}^{B} X_i^{j},\ \sigma_i^{2} = \frac{1}{B}\sum_{j=1}^{B} (X_i^{j} - \mu_i)^2
\end{equation}

Here, $\mu_i$ and $\sigma_i^{2}$ denote the running mean and running variance, respectively. $\gamma_i$ and $\beta_i$ denote the linear transformation parameters. $B$ denotes the batch size and $\epsilon$ is a small constant added for numerical stability. The proposed DAB brings two main advantages: (1) Dataset-invariant Conv layers preserve the commonality and derive homogeneous representations for different datasets, which help maintain the efficiency of network without introducing extra FLOPs and parameters for segmentation on different datasets. (2) Dataset-specific BN layers preserve the particularity and capture heterogeneous statistics across different datasets, which help alleviate the distribution shift problem during the joint optimization process.

For final predictions on different datasets, we append several dataset-specific classifiers to the output of the encoder-decoder network. Each classifier is only responsible for its corresponding dataset. Unlike the label remapping method, we do not rely on the assumption of class overlap and can exploit the out-of-the-box labels of each dataset for training.

\subsubsection{Dataset Alternation Training.}
To facilitate the collaborative optimization procedure, we introduce a dataset alternation training (DAT) mechanism to train the network. As shown in Figure \ref{fig:pipeline}, a complete iteration of DAT includes $N$ forward passes and 1 backward pass. In each forward pass, we construct the batch with samples from a single specific dataset and compute the loss separately. After all the datasets are counted, we accumulate the loss of each dataset and backpropagate the entire gradient through each dataset flow. We repeat such training procedures till network converges. To better understand the effect of DAT, we provide analysis in the following aspects. (1) Compared to backpropagating the gradients immediately after computing the loss of each dataset, DAT can reduce the training instability caused by the discrepancy of different feature distributions between two consecutive iterations. (2) Alternating between different datasets may help sufficiently leverage information from other datasets to lift the performance.




We follow common segmentation methods to use the conventional multi-class cross-entropy loss for training each dataset. Our final objective function for cross-dataset training can be formulated as below and can be minimized end-to-end.
\begin{equation}
L = -\sum_{i=1}^{N}\sum_{j=1}^{M}w^{i}y^{i}_j{\emph{l}og}(p^{i}_j) \\
\label{eq:loss}
\end{equation}
where, $N$ denotes the number of datasets, $M$ denotes the number of image pixels, $p^{i}_j$ and $y^{i}_j$ refer the predicted probability and corresponding label for the $j$-th pixel on the $i$-th dataset, respectively. $w^{i}$ denotes the loss weight and we set $w^{i}=1$ to make these loss values comparable.

\section{Experiments}

\subsection{Experimental Setting}

\subsubsection{Datasets.} We apply our CDCL method on three semantic segmentation datasets for autonomous driving: Cityscapes, BDD100K, CamVid. Dataset details are provided in the supplementary material.

\subsubsection{Implementation Details.} We use the architecture of PSPNet with the pretrained ResNet-18 / ResNet-101 on ImageNet as our baseline.
The networks are trained using stochastic gradient descent (SGD) with momentum of 0.9, weight decay of 0.0001, and batch size of 8. The initial learning rate is set to 0.01 and multiplied by $(1 - \frac{iter}{maxiter})^{0.9}$ with a polynomial decaying policy. Unless specified otherwise, we randomly crop the images into $512\times512$ for training, and use random scaling (0.5 $\sim$ 2.1) and random flipping for data augmentation. We use the standard metric of mean IoU (mIoU) to evaluate the segmentation accuracy for each dataset. More implementation details are provided in the supplementary material.

\begin{table}[t!]
\small
\centering
\begin{tabular}{l|c|c|c}
\toprule
Method           & \multicolumn{2}{|c|}{Cityscapes (\%)} & BDD100K(\%) \\
\cmidrule{2-3}
    & Val. & Test & Val.\\
\midrule
Single-dataset       &     67.52   &  67.75       &   53.88       \\
Finetuning       &    67.79    &  66.52       &   58.30   \\
Label remapping  &    66.23    &  66.39       &   58.74   \\
CDCL (Ours)             &    \textbf{72.63}    &  \textbf{71.55}       &   \textbf{60.47}  \\
\bottomrule
\end{tabular}
\centerline{\small{(a) Cityscapes + BDD100K}}
\begin{tabular}{l|c|c|c|c}
\toprule
Method     & \multicolumn{2}{|c}{Cityscapes (\%)}& \multicolumn{2}{|c}{CamVid (\%)} \\
\cmidrule{2-5}
           & Val. & Test  & Val. & Test \\
\midrule
Single-dataset   & 67.52  &  67.75 & 73.05  &70.41     \\
Finetuning   & 67.35  & 67.87  & 74.83  & 71.16\\
Label remapping    & 67.13  & 68.22  & 78.03  & 76.86 \\
CDCL (Ours)        &  \textbf{69.77}  &  \textbf{68.56} &  \textbf{78.45}  &  \textbf{77.34}\\
\bottomrule
\end{tabular}
\centerline{\small{(b) Cityscapes + CamVid}}
\caption{Performance comparisons using the same ResNet-18 backbone in the two-dataset setting.}
\label{tab:cs_bdd_cam}
\end{table}

\begin{table}[t!]
\small
\begin{center}
\begin{tabular}{l|c|c|c}
\toprule %
Method  & Cityscapes (\%) &  BDD100K (\%) & CamVid (\%) \\
\midrule
Single-dataset & 67.52 & 53.88 & 73.05\\
CDCL (Ours) & 73.17 (\textbf{+5.65}) & 60.45 (\textbf{+6.57})& 78.84 (\textbf{+5.79})\\
\bottomrule %
\end{tabular}
\end{center}
\caption{Performance comparisons using the same ResNet-18 backbone in the three-dataset setting.}
\label{tab:cs_bdd_cam}
\end{table}

\subsection{Results in Cross-Dataset Settings}
We compare the proposed CDCL algorithm with three baseline methods on multiple datasets $\{D_i\}_{i=1}^N$:
\begin{itemize}
    \item Single-dataset: Each network $net_j$ is trained on the target dataset $D_j$ separately.
    \item Finetuning: Each network $net_j$ is pretrained on a source dataset $D_i$ and then finetuned on the target dataset $D_j$.\footnote{Before training on the segmentation datasets, we use the ImageNet pretrained model as initialization in all the experiments.}
    \item Label remapping: Each network $net_j$ is trained on the combination of the target dataset $D_j$ and a subset of $D_i$ by remapping the common labels from $D_i$ to $D_j$.
\end{itemize}
For fair comparisons, our CDCL method uses the same network structure, basic loss function and training epoch as the baselines.

\subsubsection{Cityscapes + BDD100K.} In the cross-dataset setting of Cityscapes + BDD100K, the two datasets share the same semantic categories. Table \ref{tab:cs_bdd_cam} (a) shows that our CDCL method significantly outperforms all the baseline methods on both benchmarks, e.g., +5.11\% and +6.59\% mIoU over the single-dataset baseline on the validation sets of Cityscapes and BDD100K, respectively. All the three baselines produce multiple networks, each for a dataset. Differently, our method learns a unified model for all datasets without introducing extra FLOPs, and performs more efficiently without repeating the training process multiple times. We also note that label remapping even performs slightly worse than the single-dataset baseline, despite the usage of more data. We analyze that a simple combination of two datasets may cause disturbation. In contrast, our method effectively eases dataset bias and brings consistent accuracy gains on both datasets. The results demonstrate that our method can sufficiently utilize the complementarity between different datasets for performance improvement on each dataset in the case of the same label space.

\subsubsection{Cityscapes + CamVid.} In this cross-dataset setting, the two datasets have different label spaces: Cityscapes has 19 classes and CamVid has 11 classes. Table \ref{tab:cs_bdd_cam} (b) shows that our method can achieve consistent performance improvement compared to all the baseline methods, e.g., +2.25\% and +5.40\% mIoU over the single-dataset baseline on the validation sets of Cityscapes and CamVid, respectively. We note that label remapping also obtains promising results in this setting. Compared to this baseline, our method still can obtain better performance without the careful label preprocessing step. The results demonstrate the effectiveness of our CDCL method in the case of different label spaces.

\subsubsection{Cityscapes + BDD100K + CamVid.} We also evaluate our method in the three-dataset setting where Cityscapes, BDD100K and CamVid are used. Table \ref{tab:cs_bdd_cam} shows that CDCL enables over 5-point gains on all the validation sets, which validates the effectiveness of our method in the more challenging multi-dataset case.

\subsection{Ablation Study}

\subsubsection{Effect of DAB.} We use the cross-dataset setting of Cityscapes + BDD100K with the ResNet-18 backbone to examine the effect of DAB. Table \ref{tab:cs_bdd_for_dab_dat} compare four different configurations of Conv and BN layers. Using unshared BN and unshared Conv is equivalent to the single-dataset baseline. Using shared BN yields inferior performance on both benchmarks. This is because BN represents the statistics of each specific dataset, which is in line with our observation in Figure \ref{fig:conv_bn_ana}. Our DAB uses shared Conv and unshared BN and achieves the best result in these configurations. The results validate that (1) dataset-invariant Conv can obtain similar performance and reduce parameters compared to dataset-specific Conv, (2) dataset-specific BN is crucial for good performance in joint training of different datasets.

\subsubsection{Effect of DAT.} Table \ref{tab:cs_bdd_for_dab_dat} also compare the results with or without our DAT strategy in last two rows. Without DAT, the gradients are backpropagated immediately after computing the loss of each batch constructed from a single dataset. The results demonstrate that DAT can facilitate collaborative optimization and achieve improved performance on both benchmarks, e.g., +4.08\% on Cityscapes.

\subsubsection{Effect of Backbone.} We also conduct experiments using different network backbones in the cross-dataset settings of Cityscapes + BDD100K and Cityscapes + CamVid. Table \ref{tab:r101_cs_bdd_cam} shows that our CDCL method still can achieve notable improvement compared to the single-dataset baseline using a larger backbone of ResNet-101, e.g., +6.37\% mIoU on BDD100K.

\subsection{Comparisons to the State-of-the-Art Methods}
Table \ref{tab:r101_cs_sota} compares the state-of-the-art methods in terms of computation complexity (FLOPs) and segmentation performance (mIoU) on the Cityscapes test set. Through collaborative training with BDD100K, the proposed CDCL method improves PSPNet with both light-weight backbone (ResNet-18) and large backbone (ResNet-101), without increasing FLOPs in the inference phase. In addition to PSPNet, we also apply CDCL for another segmentation baseline of HANet, and also achieve improved performance with the same FLOPs. The results highlight that our method is effective, flexible and extensible for different semantic segmentation baselines and can improve the performance of state-of-the-art approaches without extra computation budgets.

\begin{table}[t!]
\small
\begin{center}
\begin{tabular}{l|c|c|c|c}
\toprule %
Conv  & BN  & DAT & Cityscapes &  BDD100K \\
\midrule
Not Shared     & Not Shared    &     \xmark         & 67.75\%     &  53.88\%   \\
Not Shared & Shared  &  \xmark            & 62.05\%     &  53.50\%   \\
Shared & Shared    &  \xmark            & 62.10\%     &  53.34\%   \\
Shared & Not Shared  &  \xmark            & 68.55\%     &  58.93\%   \\
Shared & Not Shared  &  \cmark  & \textbf{72.63\%}     & \textbf{60.47\%}   \\
\bottomrule %
\end{tabular}
\end{center}
\caption{Ablation studies on DAB and DAT with ResNet-18 on the validation sets of Cityscapes and BDD100K.}
\label{tab:cs_bdd_for_dab_dat}
\end{table}
\begin{table}[!t]
\small
\centering
\begin{tabular}{l|c|c|c}
\toprule
Method           & \multicolumn{2}{|c|}{Cityscapes (\%)}    & BDD100K (\%) \\
\cmidrule{2-3}
                 & Val. & Test & Val.\\
\midrule
Single-dataset        &  73.51     &    74.45    &    57.47       \\
CDCL (Ours)             &  \textbf{75.83}     &  \textbf{75.95}      &  \textbf{63.84}  \\
\bottomrule
\end{tabular}
\centerline{\small{(a) Cityscapes + BDD100K}}
\begin{tabular}{l|c|c|c|c}
\toprule
Method     & \multicolumn{2}{|c}{Cityscapes (\%)}& \multicolumn{2}{|c}{CamVid (\%)} \\
\cmidrule{2-5}
            & Val.       & Test           & Val.      & Test \\
\midrule
Single-dataset   & 73.51    &74.45   &  75.86   &    74.72  \\
CDCL (Ours)        &  \textbf{75.00}   &  \textbf{74.77} &  \textbf{81.15}   & \textbf{79.33} \\
\bottomrule
\end{tabular}
\centerline{\small{(b) Cityscapes + CamVid}}
\caption{Performance comparisons using the same ResNet-101 backbone in the two-dataset setting.}
\label{tab:r101_cs_bdd_cam}
\end{table}

\begin{table}[t!]
\small
\begin{center}
\scalebox{0.95}{
\begin{tabular}{l|c|c}
\toprule %
Method          & GFLOPs        & mIoU (\%)  \\
\midrule
\multicolumn{3}{c}{Current state-of-the-art results}  \\
\midrule
SegNet (Badrinarayanan et al. 2017)   &  286.0         & 56.10            \\
ENet \cite{paszke2016enet}               &  7.6           & 58.30            \\
ESPNet \cite{mehta2018espnet}            &  8.9           &  60.30           \\
ESPNetv2 \cite{mehta2019espnetv2}         &  5.4          & 65.10            \\
FCN-8s (Long et al. 2015)               &  1335.6        & 65.30            \\
ERFNet \cite{romera2017erfnet}           &  25.6          & 68.00            \\
RefineNet \cite{lin2017refinenet}        &  2102.8        & 73.60             \\
Axial-DeepLab-L \cite{wang2020axial}     &  1374.8        & 79.50             \\
HRNet \cite{wang2020deep}                & 5843.8         & 81.10             \\
BFP \cite{ding2019boundary}              &   2157.3        &  81.40  \\
DANet \cite{fu2019dual}                  &  39647.0        & 81.50   \\
OCRNet (Yuan et al. 2019)            &  5843.8        & 81.60 \\
\midrule
\multicolumn{3}{c}{Results w/o and w/ our scheme}   \\
\midrule
PSPNet (ResNet-18) \dag    &   512.8   &  67.75 \\
PSPNet (ResNet-18) (Ours)   &   512.8   &  71.55   \\
PSPNet (ResNet-18) (Ours)\ddag  & 1730.7  &   72.52 \\
\midrule
PSPNet (ResNet-101)\dag   &   2299.8   &  77.79  \\
PSPNet (ResNet-101) (Ours)        &   2299.8         &  78.74   \\
PSPNet (ResNet-101) (Ours)\ddag  &  7762.0         &  79.73 \\
\midrule
HANet (ResNet-101) (Choi et al. 2020)  &  3160.0  & 80.90 \\
HANet (ResNet-101) (Ours)               &   3160.0  &  81.62 \\
\bottomrule
\end{tabular}}
\end{center}
\caption{Comparisons with the state-of-the-art methods on the Cityscapes test set. Both PSPNet and HANet with the ResNet-101 backbone are trained by $769\times769$ input. $\dag$ refers our reproduced results. $\ddag$ refers multi-scale testing.}
\label{tab:r101_cs_sota}
\end{table}

\subsection{Comparisons to Domain Adaption Methods}
The dataset-specific BN in our DAB is similar to existing unsupervised domain adaptation methods~\cite{chang2019domain}. Both of \cite{chang2019domain} and our method tackle the domain shift problem. The main difference lies in the optimization goals and strategies. \cite{chang2019domain} aims to improve the performance on the target domain without annotations by leveraging information from the source domain. It uses a two-stage training process, i.e., first training the network on source and then finetuning the network on target with newly initialized BN layers. Differently, our goal is to improve the performance of all datasets by collaborative training. Table \ref{tab:da_cs_bdd} compares CDCL with this two-stage baseline in our supervised cross-dataset setting of Cityscapes + BDD100K, where both methods use separate BN layers for these two datasets. With freezing Conv when training on the target domain, DA can maintain the performance of the source domain and achieves better results than directly testing (37.02\% vs. 40.28\%), but performs worse than training on target (53.88\% vs. 40.28\%). With updating Conv, the performance on target can be further improved (40.28\% $\to$ 51.16\%), but fails on source (only 1.98\% on Cityscapes). This is not surprising as Conv and BN layers for Cityscapes are not jointly trained in the second stage. Our method can obtain superior performance compared to the two-stage DA baseline on both source and target domains, which verifies the non-trivial design of DAB and DAT for collaborative optimization.

\begin{table}[t!]
\small
\begin{center}
\begin{tabular}{l|c|c}
\toprule
Method & Cityscapes (\%) & BDD100K (\%) \\
\midrule
Train on C only & 67.52 & 37.02 \\
Train on B only & 45.95  & 53.88 \\
C $\rightarrow$ B, freeze Conv & 67.52 & 40.28\\
C $\rightarrow$ B, update Conv & 1.98 & 51.16\\
CDCL (Ours) & \textbf{72.63} & \textbf{60.47}\\
\bottomrule
\end{tabular}
\end{center}
\caption{Performance comparisons with domain adaptation training strategy on the validation sets of Cityscapes and BDD100K. C: Cityscapes. B: BDD100K.}
\label{tab:da_cs_bdd}
\end{table}

\begin{table}[t]
\small
\begin{center}
\begin{tabular}{l|c}
\toprule
Method & KITTI (\%) \\
\midrule\midrule
Single-dataset (Train on C only) &  51.40 \\
Finetuning (B $\to$ C) & 47.52 \\
Label remapping (C + B) & 53.79\\
CDCL (C + B) & \bf54.05\\
\midrule\midrule
Single-dataset (Train on C only) &  54.65 \\
Finetuning (B $\to$ C) & 54.16 \\
Label remapping (C + B) & 55.89\\
CDCL (C + B) & \textbf{59.05}\\
\bottomrule
\end{tabular}
\end{center}
\caption{Zero-shot domain generalization on the KITTI dataset. First group: Directly test using pretrained BN on Cityscapes. Second group: Test with precise BN \cite{wu2021rethinking} using updated mean and variance of BN on KITTI. C: Cityscapes. B: BDD100K.}
\label{tab:dg_cs_bdd}
\end{table}
\subsection{CDCL for Different Scenes} In addition to evaluations on multiple datasets from the similar autonomous driving scenes, we also conduct experiments on different scenes with a larger data distribution gap. We use the setting of Cityscapes + Pascal Context, where Cityscapes is a driving dataset and Pascal Context \cite{mottaghi2014role}
is an everyday object dataset.
Our method achieves gains of 1.87\% and 1.27\% mIoU on the validation set of Cityscapes (67.75\% vs. 69.62\%) and Pascal Context (40.56\% vs. 41.83\%) compared to the single-dataset baseline, respectively.
Although it is more usual that different datasets from similar scenes are used to boost performance in practice, we believe that CDCL can be applied to the general case of multi-dataset joint training.

\subsection{CDCL for Zero-Shot Domain Generalization} Zero-shot domain generalization (DG) aims to generalize the model on the unknown domain where training images are not available. Table \ref{tab:dg_cs_bdd} compares different baselines with our method for this challenging task. With directly testing the trained model on the unseen KITTI dataset \cite{geiger2013vision}, CDCL achieves the best result compared to all the baselines. We also apply precise BN \cite{wu2021rethinking} which updates the mean and variance of BN through forwarding the network on the entire set of KITTI. Results show that our method still outperforms all the baselines for DG, which exhibits the potentiality of CDCL for zero-shot segmentation.

\subsection{CDCL for 3D Segmentation}
Our framework can be readily extended to point cloud 3D segmentation. Point cloud 3D segmentation is another challenging task in autonomous driving, which greatly relies on a mass of annotated data to cover diverse scenes. We use two autonomous driving datasets for evaluations: SemanticKITTI \cite{behley2019semantickitti} and nuScenes \cite{caesar2020nuscenes}.
We adopt SalsaNext as the 3D segmentation baseline with the same hyper-parameter setting in (Cortinhal et al. 2020). Our CDCL method achieves 60.0\% and 67.3\% mIoU on the validation set of SemanticKITTI and nuScenes, improving the SalsaNext baseline by 1.0\% and 0.3\%, respectively. The results validate the effectiveness of CDCL for 3D segmentation and further exhibit the generality of our method.

\section{Conclusion}
In this work, we propose a unified cross-dataset collaborative learning segmentation algorithm that is capable of learning from multiple datasets. The dataset-aware block can capture heterogeneous statistics across different datasets and maintain high performance without introducing extra FLOPs. The dataset alternation training mechanism can facilitate the collaborative optimization procedure. Our method is simple, flexible and compatible with different encoder-decoder segmentation frameworks and applicable for both 2D and 3D segmentation. We consider applying our method for Transformer-based segmentation and further enhance the role in zero-shot domain generalization in future work. We also expect that CDCL can inspire new insights for more computer vision tasks beyond segmentation.
\bibliography{CDCL_in_Autonomous_Driving}
\end{document}